\documentclass[conference]{IEEEtran}
\usepackage{times}

\usepackage[numbers]{natbib}
\usepackage{multicol}
\usepackage{tabularx}
\usepackage{multirow}
\usepackage[bookmarks=true]{hyperref}
\usepackage{cite}
\usepackage{amsmath,amssymb,amsfonts}
\usepackage{algorithmic}
\usepackage{graphicx}
\usepackage{textcomp}
\usepackage{subcaption}
\usepackage[T1]{fontenc}
\usepackage{hyperref}
\usepackage{xspace}
\usepackage{xcolor}
\usepackage{graphicx}
\usepackage{amsmath}
\usepackage[textwidth=38pt]{todonotes}
\usepackage{tablefootnote}

\usepackage[normalem]{ulem}
\usepackage{caption}

\newcommand{\modelname}{{\sc ToolNet}\xspace}

\begin{document}

\title{\modelname: Using Commonsense Generalization for Predicting Tool Use for Robot Plan Synthesis}


\author{\IEEEauthorblockN{Rajas Bansal*, Shreshth Tuli*, Rohan Paul and Mausam}
\IEEEauthorblockA{Department of Computer Science and Engineering, 
Indian Institute of Technology Delhi \\
* Contributed equally}
}

\maketitle
    
\begin{abstract}
A robot working in a physical environment (like home or factory) needs to learn to use various available tools for accomplishing different tasks, for instance, a mop for cleaning and a tray for carrying objects. The number of possible tools is large and it may not be feasible to demonstrate usage of each individual tool during training. Can a robot learn commonsense knowledge and adapt to novel settings where some known tools are missing, but alternative unseen tools are present?
We present a neural model that predicts the best tool from the available objects for achieving a given declarative goal. This model is trained by user demonstrations, which we crowd-source through humans instructing a robot in a physics simulator. This dataset maintains user plans involving multi-step object interactions along with symbolic state changes. Our neural model, \modelname, combines a graph neural network to encode the current environment state, and goal-conditioned spatial attention to predict the appropriate tool. We find that providing metric and semantic properties of objects, and pre-trained object embeddings derived from a commonsense knowledge repository such as ConceptNet, significantly improves the model's ability to generalize to unseen tools. The model makes accurate and generalizable tool predictions. When compared to a graph neural network baseline, it achieves 14-27\% accuracy improvement for predicting known tools from new world scenes, and 44-67\% improvement in generalization for novel objects not encountered during training. 

\end{abstract}


\setlength{\abovedisplayskip}{3pt}
\setlength{\belowdisplayskip}{3pt}

\IEEEpeerreviewmaketitle

\section{Introduction}\label{sec:introduction}

Advances in autonomy are enabling robots to enter human-centric domains 
such as homes and factories where we envision them performing general purpose 
tasks such as transport, assembly, and clearing. 
In such domains, we expect an intelligent robot to make effective 
use of available \emph{tools}.
%
For example, a robot asked to remove many fruits from a table 
can use a \emph{tray} to efficiently perform the task. 
Similarly, it should be able to use a \emph{ramp} to navigate to an 
an elevated platform or a \emph{stick} for fetching an object 
beyond physical reach. 
In essence, the ability to use appropriate tools can guide the robot towards feasible and efficient plans.  
\begin{figure}[!t]
    \centering
    \includegraphics[width=0.96\columnwidth]{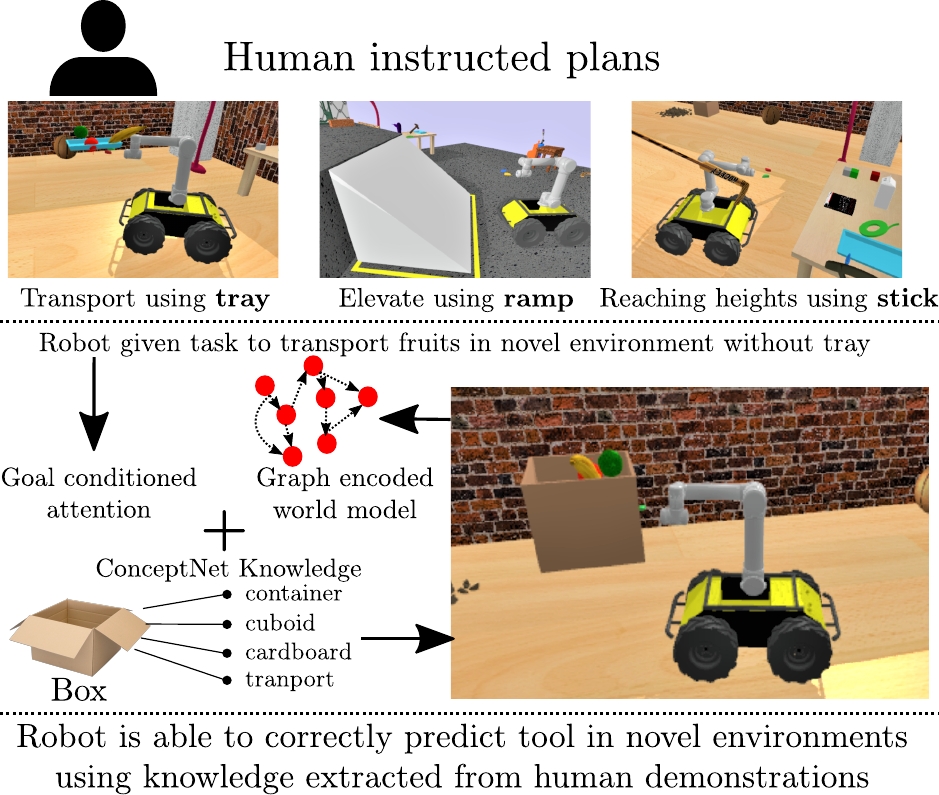}
    \caption{\footnotesize{\textbf{Overview.} Our goal is to predict the \emph{commonsense} 
    use of objects as \emph{tools} to enable a robot to perform a task. We crowd-source a data set of humans instructing a robot to perform a tasks such as transporting, fetching, clearing etc. while making use of objects such as trays, ramps, sticks as tools. A graph neural architecture predicts contextual tool use incorporating goal-conditioned attention, fusion of metric-symbolic embeddings and use of knowledge base embeddings. The learner \emph{generalizes} to novel object instances (predicting use of a "box" from  prior demonstrations of using "trays"), enabling task completion in novel contexts. 
    }}
    \label{fig:motivation}
\end{figure}

Learning the relevance of objects as tools for an intended goal is 
challenging for several reasons. 
First, the usefulness of a tool varies with context.  
For example, placing milk in the cupboard may require the robot to elevate 
itself vertically using a ramp if the milk is placed 
at a height unreachable by the robot, but if the milk is kept on a table, 
a simple tray might suffice.
Second, realistic work-spaces are typically large with an expansive space of 
possible tools and interactions. 
Acquiring data for all feasible tool objects or exploring the space of tool interactions 
is challenging for a learning algorithm. 
Ideally, an intelligent agent must be able to \emph{generalize} its knowledge and 
adapt to objects unseen at training time. 
 %
For example, knowing that trays are useful for transport tasks, 
a robot should be able to reason that a \emph{box} could be 
a useful candidate for a new transport task based on shared context and similar attributes. 

Humans possess innate commonsense knowledge
about contextual use of tools for an intended goal~(\citet{allen2019tools}). 
We hypothesize that humans possess commonsense knowledge about 
which objects could serve as tools for an intended goal. 
For example, a human actor when asked to move objects is likely to use trays, boxes, or even improvise with a new object with a flat surface. 
This work aims at enabling such commonsense generalization in a robotic agent. 
We leverage human demonstrated robot plans as a data source that elucidates commonsense knowledge about 
contextual and goal-directed tool use. 
The ability to predict useful tools for a task can guide a robot to quickly generate feasible plans. 
%
%
%

%
Our technical approach is as follows. 
We first crowd-source a dataset of human-instructed plans where a 
human teacher guides a simulated mobile manipulator to perform assembly, transport 
and fetch tasks using visible objects as tools.  
The process results in a corpus of $\approx\!1,500$ human demonstrated robot plans involving multi-step tool interactions. 
This corpus is used to 
supervise a (1-step) neural imitation learner that predicts tool applicability given the knowledge of the world state and the intended goal. 

We introduce a graph neural architecture, \textbf{\modelname}, that   
encodes both the metric and relational attributes of the world state 
as well as available taxonomic resources such as $\mathrm{ConceptNet}$~(\citet{speer2017conceptnet}). 
The \modelname model predicts tool use by learning an attention over 
entities that can potentially serve as tools. 
Implicitly, the model acquires knowledge about primitive spatial characteristics 
(typically an output of a mapping system) and semantic attributes (typically 
contained in taxonomic resources) enabling generalization to novel contexts with previously unseen objects. 
The predictions of the learned model can be utilized by an underlying symbolic planner 
while exploring feasible plans to the intended goal. 

Experimental evaluations in simulated home and factory-like environments with a mobile manipulator reveal both accurate prediction of goal-relevant tools as well as generalization to scenarios with unseen objects. 
This work contributes a step in the direction of acquiring commonsense knowledge relayed through human instruction for the purposes of attaining 
semantic goals.  
The data set and implementation is available at \url{https://github.com/reail-iitd/commonsense-task-planning}. 

\section{Related Work}\label{sec:related-works}
\textbf{Learning tool manipulation skills. } 
Learning control policies for manipulating tools has received recent attention in 
robotics. 
\citet{finn2017one} and \citet{parkinferring} learn tool manipulation policies from  
human demonstrations. 
\citet{xie2019improvisation} and \citet{yildirim2015galileo} learn physics models and 
\emph{effects} enabling goal-directed compositional use. 
\citet{liu2018physical} address the problem of learning primitive physical decomposition of tool like object through
its physical and geometric attributes enabling their human-like use. 
\citet{wu2016physics} learn physical properties of objects from unlabeled videos. 
\citet{toussaint2018differentiable} learn to compose physics tool interactions 
using a logic-based symbolic planner. 
\citet{nair2017combining} and \citet{lynch2019learning} learn to interact with objects 
in a self-supervised setup. 
Efforts such as \citet{holladay2019force}, and \citet{antunes2015robotic} 
plan tool interactions modeling contact and force interactions. 
Our paper considers the complementary problem of predicting 
\emph{which} objects may serve as tools for a given task 
while delegating the issue of tool manipulation to the 
aforementioned works. 

\textbf{Learning symbolic action sequences. }
Alternative efforts have focused on enabling robots to 
perform high-level tasks. 
\citet{puig2018virtualhome} build a data base of symbolic programs 
constituting high-level tasks in a home by using human subject 
instructing a virtual agent in a simulation environment. 
\citet{liao2019synthesizing} use the corpus to learn translations of 
domain independent task sketches to executable programs in the agent's physical context. 
\citet{shridhar2019alfred} take a similar approach by collecting natural language 
corpora describing high-level tasks and learn to associate instructions to 
spatial attention over the scene. 
Our approach draws inspiration from the above mentioned works 
in that we learn to predict tools that can be considered as  
sub-goals to guide planning for a high-level task. 
However, our problem differs in two ways. First, we explicitly model the 
physical constraints arising from a mobile manipulator interacting in the work-space.   
Second, instead of learning actions predicated on specific object instances, 
we address generalization to new object instances using 
primitive spatial and semantic characteristics. 
%

\textbf{Commonsense knowledge in instruction following. }
Acquisition of common sense knowledge has been explored for the task of 
robot instruction following. 
\citet{nyga2018grounding} present a symbolic knowledge base for procedural knowledge of tasks 
that is utilized for interpreting under specified task instructions. 
Efforts such as \citet{kho2014robo} propose a similar data base encoding common sense knowledge 
about object affordances (objects and their common locations). 
\citet{misra2016tell} use the learned model for interpreting instructions in the kitchen domain.  
\citet{chen2019enabling} present an instruction grounding model that leverages common sense taxonomic 
and affordance knowledge learned from linguistic co-associations. 
\citet{Bisk2020} consider the problem of learning physical common sense
associated with objects and interactions required to achieve tasks
from language only data sets. They study this problem in the context
of question-answering to enable synthesis of textual responses that
capture such physical knowledge.
This paper focuses on a learning common sense tool use in the context 
of following instructions that require multiple object interactions to 
attain the intended goal. 

\textbf{Synthetic Interaction Datasets. } 
Virtual environments have been used to collect human demonstrations for 
high-level tasks. 
\citet{puig2018virtualhome} introduce a knowledge base of actions required to perform activities in a virtual home environment. 
\citet{shridhar2019alfred} provide a vision-language dataset translating symbolic actions for a high-level activity to attention masks in ego-centric images. 
\citet{nyga2018cloud} curated data sets that provide a sequence \emph{How-To} 
instructions for tasks such as preparing recipes.
Others such as \citet{jain2015planit}, \citet{scalise2018natural} and \citet{mandlekar2018roboturk} present simulation environments and data sets for tasks such as learning spatial affordances, situated interaction or learning low-level motor skills. 
The present data sets possess two limitations that make them less usable 
for the learning task addressed in this work. 
First, the data sets are collected using human actors or avatars but 
do not explicitly model a robot in their environment. 
%
Second, a majority of the data sets aim at visual navigation and limited physical interaction with objects. They are 
less amenable to interactions (e.g., containment, pushing and attachment etc.) inherent in tool use. 
%

\section{Problem Setup}

%
\subsection{Robot and World Model}
We consider a mobile manipulator operating in a work space populated 
with a set of objects. 
The robot is situated in a home or factory like 
environment where the robot can affect the environment by interaction 
with objects in the scene. 
Each object is associated with a metric location and physical extent and may 
optionally possess discrete states such as $\mathrm{Open/Closed}$, $\mathrm{On/Off}$, etc. 
See Table \ref{tab:env_desc}.  
The world model is assumed to possess spatial notions 
such as $\mathrm{Near}$ or $\mathrm{Far}$. 
Objects in the world model can be supported by, contained within or 
connected with other objects (or the agent). Hence, we include  
semantic relations such as $\mathrm{OnTop}$, $\mathrm{Inside}$, $\mathrm{ConnectedTo}$ etc. 
%
%
The world state is object-centric including the metric locations of 
objects and the discrete states of symbolic attributes and relations.  

The robot possesses a set of behaviours or symbolic actions such as 
$\mathrm{Moving}$ towards an object, $\mathrm{Grasping}$, $\mathrm{Releasing/Dropping}$ or $\mathrm{Pushing}$ 
an object or  $\mathrm{Operating}$ an entity to imply actions that induce 
discrete state changes such as opening the door before exiting, turning on a switch etc. 
We assume that the robot's actions can be realized by the presence of an 
underlying controller.
%
We encode the geometric requirements for actions as symbolic pre-conditions. 
Examples include releasing an object from the gripper before grasping another, 
opening the door before trying to exit the room. 

\begin{table}
    \centering
    \begin{tabular}{|c|}
    \hline 
    \textbf{Robot Actions}\tabularnewline
    \hline 
    \begin{minipage}[t]{0.93\columnwidth}Push, Climb up/down, Open/Close, Switch on/off, Drop, Pick, Move to, Operate device,
    Clean, Release material on surface, Push until force\end{minipage}\tabularnewline
    \hline 
    \hline 
    \textbf{Object Attributes}\tabularnewline
    \hline 
    \begin{minipage}[t]{0.93\columnwidth}Grabbed/Free, Outside/Inside, On/Off, Open/Close, Sticky/Not Sticky, Dirty/Clean, Welded/Not Welded, Drilled/Not Drilled, Driven/Not Driven, Cut/Not Cut, Painted/Not Painted\end{minipage}\tabularnewline
    \hline 
    \hline
    \textbf{Semantic Relations}\tabularnewline
    \hline 
    On top, Inside, Connected to, Near\tabularnewline
    \hline 
    \hline
    \textbf{Metric Properties}\tabularnewline
    \hline 
    Position, Orientation, Size\tabularnewline
    \hline 
    \hline
    \textbf{Home Objects}\tabularnewline
    \hline 
    \begin{minipage}[t]{0.93\columnwidth}floor$^{1}$, wall, fridge$^{123}$, cupboard$^{123}$, tables$^{1}$, couch$^{1}$, \textbf{big-tray}$^{1}$, \textbf{tray}$^{1}$, \textbf{book}$^{1}$,
    paper, cubes, light switch$^{4}$, bottle, \textbf{box}$^{2}$, fruits, \textbf{chair}$^{15}$, \textbf{stick}, dumpster$^{2}$,
    milk carton, shelf$^{1}$, \textbf{glue}$^{6}$, \textbf{tape}$^{6}$, \textbf{stool}$^{15}$\tablefootnote{Stool/ladder are proxies for raising the height of the robot}, \textbf{mop}$^{8}$, \textbf{sponge}$^{8}$, \textbf{vacuum}$^{8}$, dirt$^{7}$, door$^{2}$\end{minipage}\tabularnewline
    \hline 
    \hline 
    \textbf{Factory Objects}\tabularnewline
    \hline 
    \begin{minipage}[t]{0.93\columnwidth}floor$^{1}$, wall, \textbf{ramp}, worktable$^{1}$, \textbf{tray}$^{1}$, \textbf{box}$^{2}$, crates$^{1}$, \textbf{stick}, long-shelf$^{1}$,
    \textbf{lift}$^{1}$, cupboard$^{123}$, \textbf{drill}$^{4}$, \textbf{hammer}$^{49}$, \textbf{ladder}$^{5}$, \textbf{trolley}$^{2}$, \textbf{brick}, \textbf{blow dryer}$^{48}$,
    \textbf{spraypaint}$^{4}$, \textbf{welder}$^{4}$, generator$^{4}$, \textbf{gasoline}, \textbf{coal}, \textbf{toolbox}$^{2}$, \textbf{wood cutter}$^{4}$,
    \textbf{3D printer}$^{4}$\tablefootnote{3D printer is a proxy for known but unobserved objects}, screw$^{9}$, nail$^{9}$, \textbf{screwdriver}$^{49}$, wood, platform$^{1}$, oil$^{7}$, water$^{7}$,
    board, \textbf{mop}$^{8}$, paper, \textbf{glue}$^{6}$, \textbf{tape}$^{6}$, assembly station, spare parts, \textbf{stool}$^{15}$\end{minipage}\tabularnewline
    \hline 
    \end{tabular}
    \caption{\footnotesize{\textbf{Domain Representation. } Robot symbolic actions, semantic attributes, relations to describe the world state and objects populating the scene in Home and Factory Domains. Legend:- $^{1}$:~surface, $^{2}$:~can open/close, $^{3}$:~container, $^{4}$:~can operate, $^{5}$:~can climb, $^{6}$:~can apply, $^{7}$:~can be cleaned, $^{8}$:~cleaning agent, $^{9}$:~can 3D print. Bold objects can be used as
tools.}}
    \label{tab:env_desc}
\end{table}

\subsection{Semantic Goals and Interactions} 
The robot's goal is to perform tasks such as transporting or delivering 
objects to appropriate destinations, making an assembly, clearing or packing items or performing 
abstract tasks such as illuminating or cleaning the room. 
We assume that the robot is instructed by providing \emph{declarative} goals.     
For example, the task of moving all fruits must be on the kitchen table 
can be modeled as a set intended constraints between the objects of interaction.  
%
%
Finally, the robot must synthesize a plan of executable actions to 
satisfy the goal constraints. 
The presence of a rich space of interactions gives rise to 
plans with multiple interactions between objects. 
For example, "packing items into a basket and carrying the basket to the goal region", 
"using a stick to fetch and drop an object beyond reach into a box", 
"using a ramp/stool to elevate itself to fetch an object".

\subsection{Predicting Generalized Tool Use}
We assume that the robot is primed with a set of primitive symbolic actions 
but lacks knowledge about how object characteristics can facilitate their use 
as in attaining high-level goals. 
Hence, the robot cannot predict\footnote{Except by discovering via explicit simulation which may be infeasible or intractable in large planning domains.}  
the use of tray-like objects in transportation tasks, or the use of a stick to fetch an object at a distance. 
%
%
%
%
Indeed, it is such commonsense association between semantic goals 
and use of objects as tools that we seek to learn. 
Thus, as a step towards finding a satisficing plan to the goal, this 
work investigates the intermediate problem of learning to predicting 
the best tool to use to achieve the given goal. 
%

Formally, let $\mathcal{O}$ denote the set of objects present in the work-space. 
Let $\mathcal{S}$ denote the world state consisting of 
metric locations of the objects (including the robot) as well as 
the set of expressed semantic relationships between entities. 
Next, we denote the goal provided to the robot as 
$\Lambda_{g}$ composed of linguistic description of the semantic constraints 
that must be satisfied by the agent (for example, "place books on the cupboard"). 
%
%
Let $\tau$ denote the set of tool objects that the robot can use in its plan\footnote{Note that only movable objects in the scene are 
considered as potential tools. Hence, $\tau  \subseteq  \mathcal{O}$.}, denoted in bold in Table~\ref{tab:env_desc}.  
%
%
%
Our goal is to predict a tool useful for the goal in 
the context of the current world state. To achieve this, our model learns to output a likelihood 
$p(t \in \tau \mid \mathcal{S}, \Lambda_{g})$. 
%
Online, the robot may encounter \emph{unseen} objects in its environment. 
Hence, we consider the \emph{open world} setup where the 
robot must generalize its knowledge to reason over 
\emph{novel} object instances, that may be unseen in training. 
%

\subsection{Learning from Human Teaching} 
We assume the presence of human teachers who can guide the robot to perform a range of 
tasks in the environment. 
Human guidance is in the form of a sequence of symbolic actions for the robot to execute 
in order to complete the desired task.    
We assume that the human teachers are cooperative and instruct the robot to utilize  
tools in order to efficiently find a feasible plan to the stated goal. 
The data set of human instructed robot plans 
elucidates our common sense knowledge about contextual tool use.
%
%
%
Note that plan variations can occur between human teachers where different objects may be 
utilized as tools in pursuit of similar goals.   
We use the demonstration data set of plans to supervise a (1-step) imitation learner 
with a range of tool use occurrences in varied world context and a diverse set of goals in the environment. 
Online, the learned model predicts the relevance likelihood of tools given novel contexts and goals 
which can be utilized by the robot to find a feasible plan. 

\section{Dataset Creation From Human Demonstrations}
\label{sec:data_collection}

\begin{figure}
    \centering
    \includegraphics[width=\linewidth]{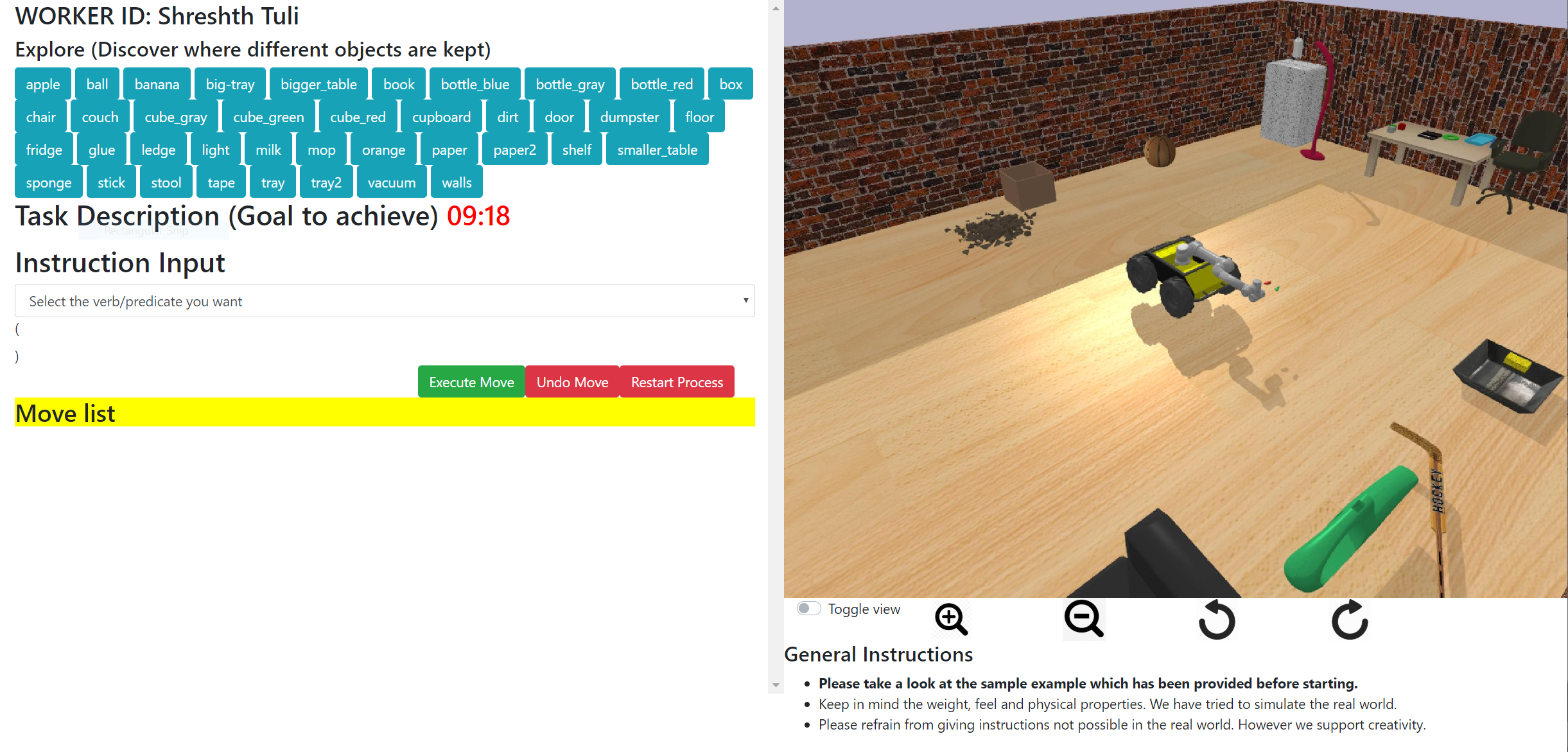}
    \caption{
    \footnotesize{
    \textbf{Data collection platform.} 
    The human teacher instructs the virtual mobile manipulator robot 
    by specifying symbolic actions. The human-instructed plan is then simulated and visualized using the interface.}
    }
    \label{fig:simulator}
\end{figure}

Our goal is to learn common sense knowledge about tool usage from 
human teachers instructing the robot to attain declarative goals. 
We proceed by creating a physics simulation environment to model a mobile manipulator\footnote{a Universal Robotics Manipulator (UR5) mounted on a Husky mobile base} capable of executing symbolic actions for a diverse range of tasks.  
We then use the environment to collect a data set of human demonstrations that instruct the robot to perform tasks in the environment.  

\subsection{Virtual Environment and User Interaction}

A simulation environment (based on \emph{Bullet} physics engine by~\citet{coumans2016pybullet}) was used to 
encode a home-like and a factory-like domain. 
A virtual mobile manipulator could pursue the following categories of semantic goals: (a) 
\emph{transporting} objects from one region to another (including space on top of or inside 
other objects), (b) \emph{fetching} objects where the robot must reach, grasp and return with, and (c) \emph{inducing state changes} such as illuminating the room or removing dirt on the floor.  
%
%
The robot's interactions were implemented with a 
low-level motion planner with a set of encoded discrete conditions 
such as moving close to an object before manipulating etc. arising 
due to physical constraints of the robot. 
The set of abstract interactions such as attachment, operating a tool or specifics of grasping 
were encoded symbolically as the establishment or release of constraints. 
The effects of actions such as pushing, moving and contact with a stick-like object 
were simulated and propagated to the next time step. 
Two domains of home and factory were implemented with objects as states in Table~\ref{tab:env_desc}. 
The objects in the domains were derived from real-world home and factory scenes and diverse object types that span Facebook
Replica Dataset (\citet{straub2019replica}) and YCB object dataset (\citet{calli2017ycb}).
For each domain, 8 different goals were given to human instructors as stated in Table~\ref{tab:dataset}.
A lexical parser was built to convert goal specification to constraints on the state of the simulator.
Moreover, 10 different scenes for each domain were created by randomly positioning different objects 
based on likely semantic placements. 
For instance, fruits could be placed $\mathrm{OnTop}$ of table or $\mathrm{Inside}$ the fridge.

A human instructor could interact with the robot agent by selecting a goal and 
specifying a sequence of symbolic action to execute. 
The specified plan was then simulated showing the robot interacting 
with objects and changing the world state. 
The human subjects are encouraged to instruct the robot such that the task 
is completed as quickly as possible, making use of available tools in the environment. 
%
%
Figure \ref{fig:simulator} illustrates the user interface used for data collection in simulation. 
%

\begin{table}[t]
    \centering
    \resizebox{\columnwidth}{!}{\begin{tabular}{|c|c|c|c|c|c|}
    \hline 
    Goal ID & Goal Text & Actions & Time & Interacted Objects & Tools used\tabularnewline
    \hline 
    \hline 
    \multicolumn{6}{|c}{Home Domain}\tabularnewline
    \hline 
    \hline 
    1 & Place milk in fridge & 5.15$\pm$2.14  & 40.03$\pm$0.04  & 3.11$\pm$1.04 & 0.62$\pm$0.70\tabularnewline
    \hline 
    2 & Put fruits in cupboard & 5.15$\pm$1.72  & 48.04$\pm$0.03 & 4.69$\pm$1.10 & 0.60$\pm$0.57\tabularnewline
    \hline 
    3 & Remove dirt from floor & 4.26$\pm$1.10  & 18.21$\pm$0.01 & 3.33$\pm$0.91 & 1.43$\pm$0.51\tabularnewline
    \hline 
    4 & Stick paper to wall & 4.31$\pm$2.19 & 41.29$\pm$0.02 & 4.41$\pm$1.17 & 1.38$\pm$0.51\tabularnewline
    \hline 
    5 & Put cubes in box & 6.47$\pm$2.61 & 66.80$\pm$0.06 & 6.25$\pm$1.18 & 0.79$\pm$0.67\tabularnewline
    \hline 
    6 & Place bottles in dumpster & 7.46$\pm$2.02 & 313.92$\pm$0.12 & 6.99$\pm$1.10 & 1.24$\pm$0.72\tabularnewline
    \hline 
    7 & Place a weight on paper & 2.07$\pm$1.59  & 33.06$\pm$0.04 & 2.69$\pm$0.95 & 0.70$\pm$0.63\tabularnewline
    \hline 
    8 & Illuminate the room & 3.17$\pm$1.85  & 13.29$\pm$0.01 & 1.77$\pm$0.62 & 0.64$\pm$0.50\tabularnewline
    \hline 
    \hline 
    \multicolumn{6}{|c}{Factory Domain}\tabularnewline
    \hline 
    \hline 
    1 & Stack crates on platform & 12.59$\pm$2.42 & 745.55$\pm$0.46 & 5.71$\pm$0.99 & 1.37$\pm$0.75\tabularnewline
    \hline 
    2 & Stick paper to wall & 8.79$\pm$2.80 & 495.48$\pm$0.11 & 4.74$\pm$1.35 & 1.76$\pm$0.87\tabularnewline
    \hline 
    3 & Fix board on wall & 12.35$\pm$3.41 & 638.54$\pm$0.29 & 4.73$\pm$0.94 & 2.08$\pm$0.87\tabularnewline
    \hline 
    4 & Turning the generator on & 9.10$\pm$2.03 & 449.01$\pm$0.09 & 3.15$\pm$0.90 & 0.74$\pm$0.72\tabularnewline
    \hline 
    5 & Assemble \& paint parts & 22.60$\pm$5.31 & 1027.33$\pm$0.83 & 7.31$\pm$1.38 & 2.82$\pm$0.87\tabularnewline
    \hline 
    6 & Move tools to workbench & 8.47$\pm$1.80 & 514.79$\pm$0.21 & 4.54$\pm$0.99 & 0.73$\pm$0.67\tabularnewline
    \hline 
    7 & Clean spilled water & 4.73$\pm$1.94 & 247.90$\pm$0.31 & 2.46$\pm$0.56 & 1.05$\pm$0.22\tabularnewline
    \hline 
    8 & Clean spilled oil & 5.11$\pm$1.72 & 361.17$\pm$0.37 & 2.44$\pm$0.57 & 1.01$\pm$0.10\tabularnewline
    \hline 
    \end{tabular}}
    \caption{\footnotesize{\textbf{Dataset characteristics. } The average number of actions, the mean object interactions and tools used for plans with associated goals for the human demonstrated robot plans collected in the home and factory domains. }}
    \label{tab:dataset}
\end{table}

\begin{figure}[!b]
    \centering
    \begin{subfigure}{0.49\linewidth}
        \centering
        \includegraphics[width=\linewidth]{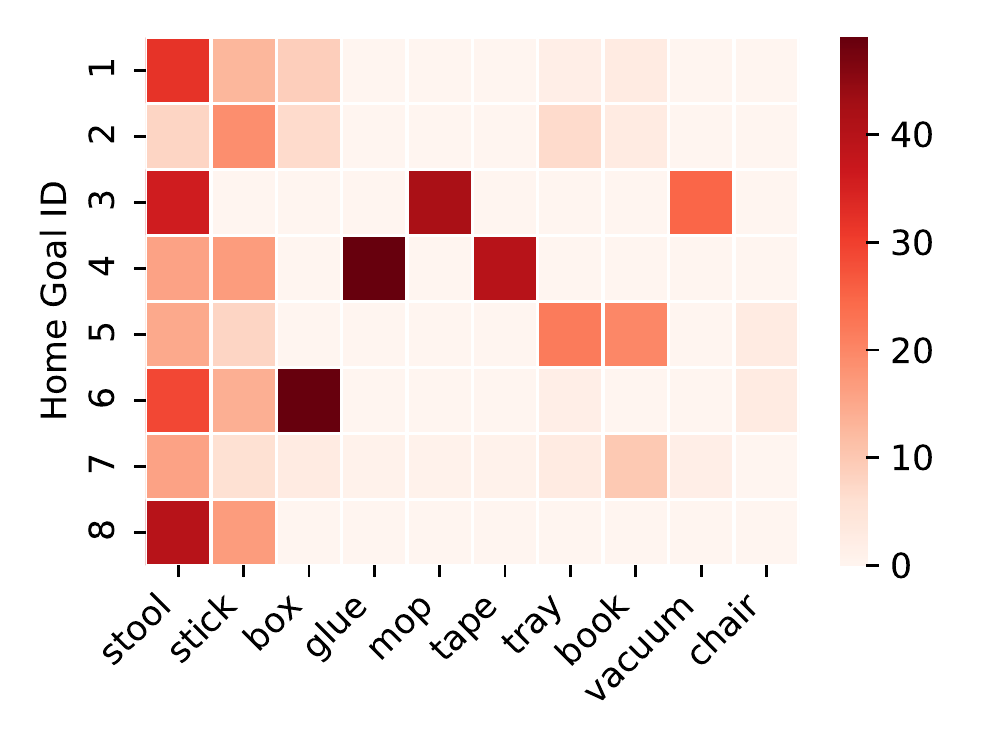}
        \caption{Home domain}
        \label{fig:home_goal_tools}
    \end{subfigure}
    \begin{subfigure}{0.49\linewidth}
        \centering
        \includegraphics[width=\linewidth]{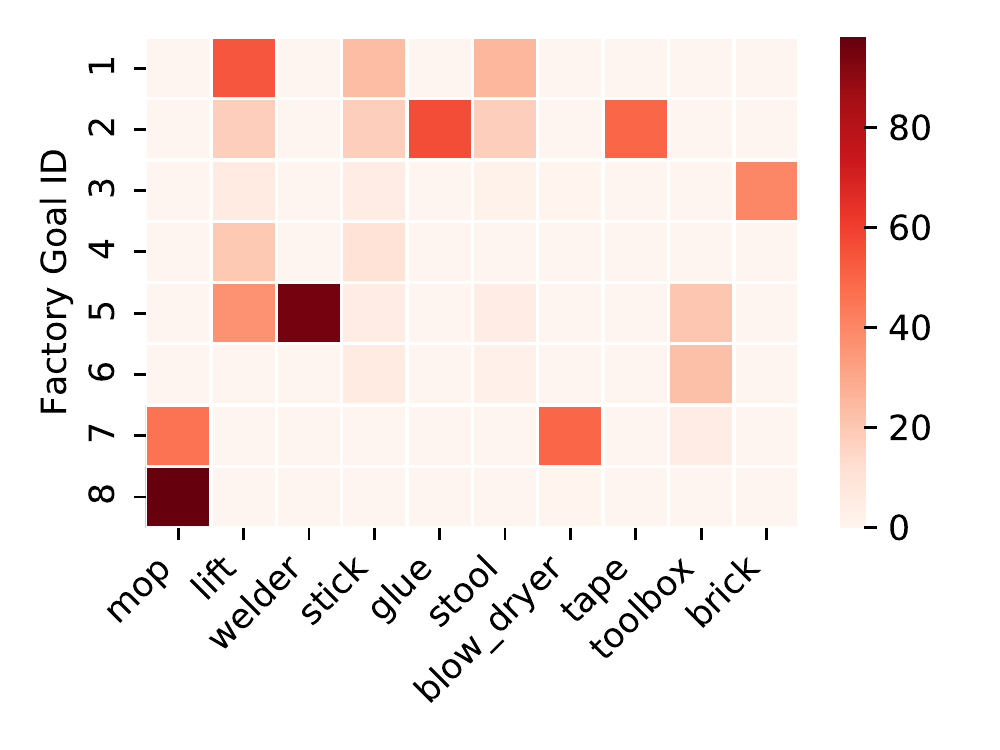}
        \caption{Factory domain}
        \label{fig:factory_goal_tools}
    \end{subfigure}
    \caption{\footnotesize{\textbf{Diversity of tool usage in human demonstrated plans. } The distribution of tool usage count of top 10 tools with associated goals in the dataset. Human instructed plans show diversity in 
    the use of tools for the robot to attain the intended goals.}}
    \label{fig:tool_with_goals}
\end{figure}

\subsection{Dataset Characteristics}
%

The dataset was collected from human instructors by presenting different $\mathrm{goal,scene}$ pairs.
For each domain, every goal was combined with each scene to give $8\times10=80$ such pairs.
For each pair, 8 to 12 plans were collected based on the level of complexity and possible plan diversity.
From 12 instructors, a complete set of $\mathrm{708}$ plans for home and $\mathrm{784}$ plans for factory domain were collected.
The original set of human demonstrated robot plans provides a limited set of environment contexts in which the plan 
are feasible. 
We augment the original set of demonstrations by exploring successful plans in perturbed environment contexts. 
%
%
Two plan augmentation strategies were used. Firstly, the plans that were successful for a goal in a 
world scene were tested on other scenes and were added to the dataset if on execution they lead
to the goal state in the simulated environment.
Secondly, we created more plans by randomly removing upto 5 objects that
are neither in the goal description nor interacted with in the plan.
After augmentation, the training datasets consist of $\mathrm{3540}$ 
and $\mathrm{3920}$ plans for home and factory, respectively.


Our dataset contains activities with several examples. 
Table~\ref{tab:dataset} analyzes the plan diversity and shows the variation of 
actions, total plan simulation time, number of interacted 
objects and tools with goals in each domain. 
The number of objects interacted with and number of tools used varies greatly as
goals change.
Plans corresponding to simpler goals like ``switching off light'' are much shorter in terms of 
plan length and execution time compared to goals like ``assemble and paint parts''.
Figure~\ref{fig:tool_with_goals} shows the
distribution of tool usage with different goals.
The frequency distribution of tool usage varies significantly with goals as different tools
are required to perform different activities. 
Different tools were used by human demonstrators for the same goal. 
Furthermore, placement of objects changes the tool usage distribution. For instance, a tray closer to
fruits is more likely to be chosen by a human instructor compared to a box far away.

\section{Learning to Predict Tool Use}\label{sec:model}

\begin{figure*}[!t]
    \centering
    \includegraphics[width=0.95\textwidth]{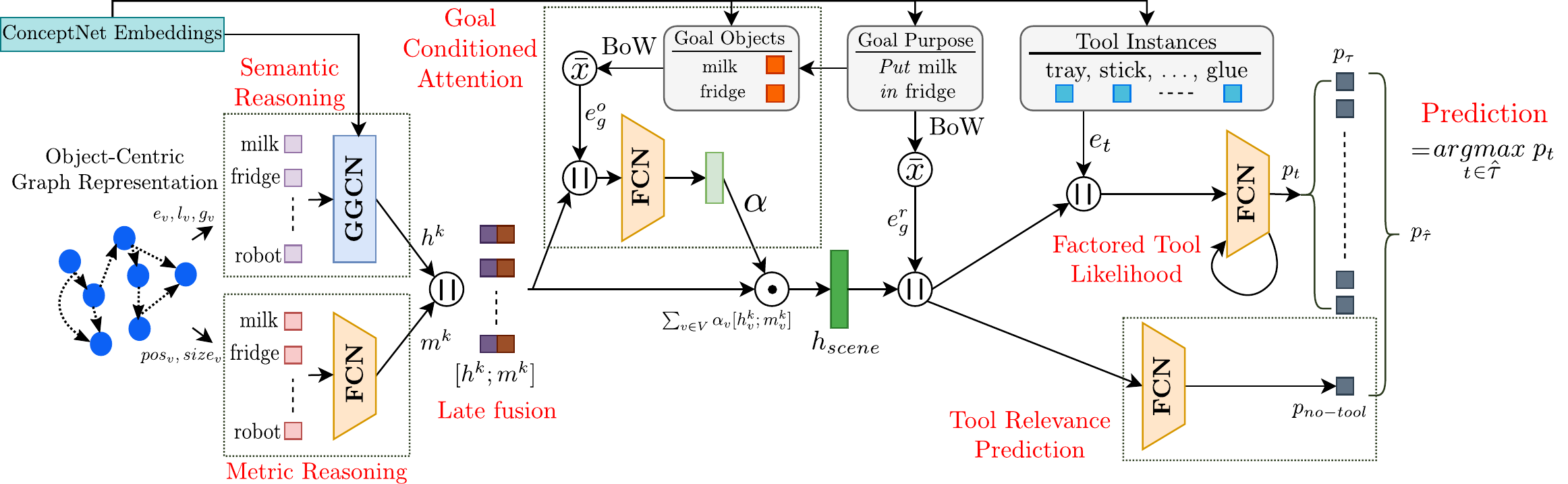}
    \caption{\footnotesize{\textbf{Overview of the \textbf{\modelname} model.} The object-centric graph representation of the agent's environment is encoded by late fusion of metric and semantic encodings. Goal conditioned attention is applied to this representation, which with goal and tool embeddings is used to predict tool likelihoods. }}
    \label{fig:complete_model}
\end{figure*}
We assume an object-centric environment representation and 
model the robot's world state as a graph expressing object attributes and relations.  
Given the current world state $\mathcal{S}$ (as a graph) and the 
goal description $\Lambda_{g}$, our neural model estimates the likelihood 
$p(t \mid \mathcal{S}, \Lambda_{g})$ over candidate tool objects $t \in \tau$ 
in the environment.  
We build on the ResActGraph model by \citet{liao2019synthesizing} as the baseline 
and extend the model to our problem setup. 
Figure~\ref{fig:complete_model} presents the final \modelname model. 

\subsection{Graph-structured World Representation (GGCN)}
\label{sec:ggcn}

We encode the robot's world state $\mathcal{S}$ in the form of an object-centric graph. 
The graph $G = (V,R)$ consists of object 
instances as the vertex set $V \subseteq \mathcal{O}$ and semantic relations as the edge set 
$R \subseteq V \times V$. 
Each node $v \in V$ indicates the object instance of the scene 
and has a pre-trained FastText embedding $e_v$ (\citet{joulin2016bag}), its semantic states $l_v$, 
and metric properties including position $pos_v$ and size $size_v$. 
Note that $V$ includes a node for the agent. The edge $r \in R$ encodes semantic relations 
between two objects $v_1$ and $v_2$ as $r(v_1, v_2) \in R$. 
Each node also has a bit, $g_v$ which is set to 1 if the object is mentioned in the goal and 0 otherwise.

The node features along with relations are used to obtain vector embeddings for each object using Gated Graph Convolution Networks (GGCNs) (\citet{liao2019synthesizing}) as described below.
The initial embedding of each node is a concatenation of all features mentioned above, i.e $[e_v;l_v;pos_v;size_v;g_v]$. The hidden state of a node, $h^0_v$ is initialized as a lower dimensional projection of the initial embedding for the object:
\begin{equation}
    h_v^{0} = tanh(W_{init}[e_v;l_v;pos_v;size_v;g_v]),
\end{equation}
where $W_{init}$ are learnable weights.
At each propagation step $s$, each node embedding is created using the hidden states of its neighbours $v' \in N(v)$ at propagation step $s-1$, so applying graph convolutions,
\begin{equation}
    x^s_v = \sum_{j \in R}\sum_{v' \in N(v)} W_{p_j}h^{s-1}_{v'}.
\end{equation}

After aggregating this information, the gating stage of the GGCN is realized using a Gated Recurrent Unit (GRU) described 
as follows,
\begin{equation*}
    z^s_v = \sigma(W_z[x^s_v;h^{s-1}_v] + b_z),
\end{equation*}
\begin{equation*}
    r^s_v = \sigma(W_r[x^s_v;h^{s-1}_v] + b_r),
\end{equation*}
\begin{equation*}
    \hat{h^s_v} = tanh(W_h[x^s_v;r^s_v \odot h^{s-1}_v] + b_h),
\end{equation*}
\begin{equation*}
    h^s_v = (1 - z^s_v) \odot h^{s-1}_v + z^s_v \odot \hat{h^s_v}.
\end{equation*}
We use k = 2 propagation steps of gated graph convolutions in our model.
This results in a hidden state vector $h^k_v$ for each node $v$ in the graph which 
aggregates all information about the corresponding object. 
The embedded node vectors are added together to get the scene representation, i.e
\begin{equation}
    h_{scene} = \sum_{v \in V} h^k_v.
    \label{eq:ggcn_graph_embedding}
\end{equation}

The goal is embedded through a \emph{bag-of-words} (BoW) using FastText embeddings ($f_w$ is the embedding of the word $w$) of the goal text representing the goal $\Lambda_g$. Thus the goal embedding ($e_g^r$), is
\begin{equation}
    e^r_g = \frac{1}{|\Lambda_G|} \times  \sum_{w \in \Lambda_G} f_w.
\end{equation}
We augment the tool set $\tau$ with \emph{no-tool} category, to give $\hat{\tau}$, to account for the case where a tool is not required to achieve the goal.
In the base model, the scene and the goal encoding are used to predict a distribution over $\hat{\tau}$. Hence,
\begin{equation}
    p_{\hat{\tau}} = \sigma(W[h_{scene};e_g^r]).
\end{equation}

\subsection{Fusion of Metric and Semantic Attributes (+Metric)}
When we have two semantically different feature sets, symbolic and metric, passing them through separate networks allows our model to exploit them independently. Thus, unlike the baseline GGCN model, we handle the semantic attributes using graph convolutions (GGCN) and metric attributes using Fully Connected Network (FCN). We then combine the two representations to form the scene embedding (the late fusion as shown in the figure). This late fusion allows the downstream prediction to give more emphasis to the separated feature, which may be lost in early fusion (\citet{lu2014video}).
Late fusion of the metric properties along with the hidden states of objects obtained
through the GCN ensures that information like object position and size are available when predicting tool likelihoods. 

The metric properties are thus encoded separately using a Parameterized ReLU (PReLU) layer $m^0_v = PReLU(W^0_{metric}[e_v;pos_v;size_v)$. At $k^{th}$ layer, 
\begin{equation}
    m^k_v = PReLU(W^k_{metric}[m^{k-1}_v]).
\end{equation}
Now, the hidden state used to embed the scene is $[h^k_v;m^k_v]$ instead of just $h^k_v$. Hence, $h_{scene} = \sum_{v \in V} [h^k_v;m^k_v]$.

\subsection{Goal-Conditioned Attention (+Attn)}
Our current encoding of the environment is independent of the goal that needs to be completed by the tool.
As common work spaces have large number of objects, we estimate a local context using goal specifications.
We create a goal-conditioned view of the environment by learning an attention over the scene objects using the goal information.
This helps the robot to ignore the myriad of "distraction objects" present in the scene.
%
The attended scene embedding is constructed using Bahadanau style attention (\citet{bahdanau2014neural}), where attention weights are calculated from the hidden state $[h^k_v;m^k_v]$ as well as the \emph{bag-of-words} embeddings, $e^o_g$, for goal objects, $\Lambda^o_g$ (the goal conditioned attention module in figure~\ref{fig:complete_model}). Thus, we define
\begin{equation}
    h_{scene} = \sum_{v \in V} \alpha_v [h^k_v;m^k_v] \mathrm{\ where,}
\end{equation}
\begin{equation}
\label{eqn:attn_scene}
    e^o_g = \frac{1}{|\Lambda^o_G|}\! \cdot\! \sum_{w \in \Lambda^o_G} f_w \mathrm{\,,and\,} \alpha_v = softmax(W[h^k_v;m^k_v;e^o_g]).
\end{equation}
    

\subsection{Factored Likelihood over Tool Instances (+L)}
The model discussed till now does not possess the ability to generalize to objects not seen during training time.
%
%
In order to allow the model to generalize to unseen tools, instead of prediction over the pre-defined tool set $\hat{\tau}$, we allow the model to predict a likelihood score of a tool $t$ (which may not be present in any of the scenes in the training set) to be used as a tool using its FastText embedding ($e_t$). This recurrence is shown in the factored tool likelihood module in Figure~\ref{fig:complete_model}. For the \emph{no-tool} case, an embedding consisting of all zeros is used. Likelihood of each tool is computed for each $t$ using,
\begin{equation}
    p_{t} = \sigma(W[e_t;h_{scene};e^r_g])\ \forall\ t \in \hat{\tau}.
\end{equation}

\subsection{Incorporating Tool Relevance/Non-relevance (+NT)}
We observe that not using any tool (\emph{no-tool}) to complete a goal is possible for a large set of scene-goal pairs and it is difficult for the model to learn in this skewed class distribution. Thus we factor the problem into two parts: (a) predicting if a tool will be used (the tool relevance module in the figure), and (b) predicting which tool will be used, given a tool needs to be used (the factored tool likelihood module in the figure). Thus (b) predicts over the set $\tau$, which gives
\begin{equation}
    p_{\tau} = \operatornamewithlimits{||}_{t\in \tau} \sigma(W[e_t;h_{scene};e^r_g]).
\end{equation}
Prediction of (a) is done using,
\begin{equation}
    p_{no-tool} = \sigma(W[h_{scene};e_g^r]).
\end{equation}
Combining the two conditional tool likelihoods, we get
\begin{equation}
    p_{\hat{\tau}} = (1 - p_{no-tool}) \cdot [p_{\tau};0] + p_{no-tool} \cdot [0;\ldots0;1].
\end{equation}

\subsection{Incorporating Learned Knowledge base Embeddings (+C)}
Pretrained $\mathrm{ConceptNet}$ Numberbatch embeddings~(~\citet{conceptnet-github}) are used as a source of relational knowledge about objects like relative sizes, relations like \emph{similar to} and \emph{capable of}. 
It is built using an ensemble that combines data from $\mathrm{ConceptNet}$, word2vec, GloVe, and OpenSubtitles 2016, using a variation on retrofitting with $\mathrm{ConceptNet}$ knowledge graph containing semantic information.
Retrofitting ensures objects of  similar "type" to have closer vector representations.
$\mathrm{ConceptNet}$ Numberbatch provides a richer semantic space for generalization compared to FastText embeddings.
It allows our model to incorporate a rich space of semantic properties and relations to extrapolate to other tools and tool types from tool use observed in human demonstrated plans.  

\subsection{Training Loss}

The loss used to train all models is Binary Cross-Entropy with each $v \in \hat{\tau}$ acting as a class. The class label, $y^i_j$ is assigned 1 if it is used in a given demonstration, $i$ and 0 otherwise. Thus, 
the loss function is defined as

\begin{equation}
    \mathcal{L} = - \sum_i \sum_j y^i_j \log(p(y^i_j)) + (1 - y^i_j) \log(1 - p(y^i_j)).
\end{equation}

We use categorical weights based on plan execution time to encourage shorter plans. However, the knowledge of the time taken for different plans has not been injected into the model.
In order to make this notion explicit to the model we use loss weighting (\emph{+W}) such that,
\begin{equation}
    \mathcal{L} = - \sum_i \alpha_i  \sum_j y^i_j \log(p(y^i_j)) + (1 - y^i_j) \log(1 - p(y^i_j)),
\end{equation}
where, $\alpha_i$ is a high for optimal plans (shortest among human demonstrations) and low otherwise.

\section{Results}\label{sec:results}


Our experiments answer the following questions. (1) What is the performance of our complete \modelname architecture compared to baseline models for the task of tool prediction? (2) How robust is \modelname when tested in a variety of generalization scenarios such as unseen tools and other objects? (3) What is the incremental contribution of each model improvement in \modelname, and where does it help?

\subsection{Evaluation Setup}

We test \modelname's tool prediction capabilities in two settings. In the first setting, we use the dataset as described in Section~\ref{sec:data_collection} and split it 
according to the scene instance. We use augmented data corresponding to 9 scenes as the training set
and split data for the remaining scene instance equally to form validation and testing sets.
%
We use accuracy as our performance measure. Here, a tool prediction is deemed correct if the predicted tool is used in at least one of the various annotated plans for the $(\rm{goal,scene})$ pair and incorrect otherwise.

To test model's generalization abilities, we generate novel  $(\rm{goal,scene})$ pairs by sampling unseen object locations and 
object replacement based on contexts specified in Table~\ref{tab:generalization_testcases}.
It describes five types of generalization test-cases.
In Type I,  we change an object's position to test whether, of the multiple possible choices, the model predicts the tool that is closer to the goal object.  For example, we replace the positions of \emph{tray} and \emph{box} for the goal "place fruits
in the cupboard" to check if the model predicts the tool closer to the fruits.
In Type II, we remove the tool predicted by the model for all plans in the training set. 
This verifies whether the model predicts a reasonable alternative if the most likely tool is absent from the scene.
In Type III, we replace a tool with an alternate tool that is unseen at training time. For example, \emph{box} is replaced with \emph{bucket} and \emph{stool} with \emph{ladder}. We extract these object/tool replacements from the $\rm{ConceptNet}$ graph.
Type IV scenes replace a tool with random objects unrelated to the task (\emph{stool} to \emph{headphone}) or with tools used for a different task. The goal is to evaluate the model's ability to predict an alternative tool relevant for the 
task or estimate the absence of a relevant tool. 
%
Finally, we also replace goal objects to check if the model predicts alternate tools or not based on
new object's metric/semantic properties like size/position. For example, to transport an apple, a \emph{tray} might work. 
However, to transport larger object like pillow, a \emph{box} would be needed.
We make these changes on the existing annotated data points and call it 
the GenTest data-set, which consists of $\mathrm{1,406}$ scenes.

Since predicting tool is a new problem, no existing algorithms exist for it. We compare against the basic GGCN model of Section~\ref{sec:ggcn} as our baseline model, since its encoder incorporates technical ideas from recent imitation learning works (\citet{shridhar2019alfred, liao2019synthesizing}) on action prediction.

\subsection{Results}

\begin{table}
    \centering
    \resizebox{\columnwidth}{!}{
    \begin{tabular}{|c|c|c|}
    \hline 
    Type Number & Generalization context & Examples\tabularnewline
    \hline 
    \hline 
     I &  \begin{minipage}[t]{0.4\columnwidth} Change goal object location\end{minipage} &  \begin{minipage}[t]{0.4\columnwidth}Replace tray positions, Place milk carton on table instead of on top of fridge \end{minipage}\tabularnewline
    \hline 
    II & \begin{minipage}[t]{0.4\columnwidth}Remove maximum likelihood tool \vspace{0.1cm} \end{minipage} & \begin{minipage}[t]{0.4\columnwidth}Remove tray, mop, glue, box, wood\end{minipage}\tabularnewline
    \hline 
    III & \begin{minipage}[t]{0.4\columnwidth}Tool to alternate tool replacement\end{minipage} & \begin{minipage}[t]{0.4\columnwidth}Box $\rightarrow$ crate, basket; stool $\rightarrow$ seat,
    step-ladder; toolbox $\rightarrow$ box, bucket\end{minipage}\tabularnewline
    \hline 
    IV & \begin{minipage}[t]{0.4\columnwidth}Tool to non alternate object replacement \end{minipage}& \begin{minipage}[t]{0.4\columnwidth}Stool $\rightarrow$ headphone; Lift $\rightarrow$ headphone\end{minipage}\tabularnewline
    \hline 
    V & \begin{minipage}[t]{0.4\columnwidth}Goal object to another goal object replacement \end{minipage}& \begin{minipage}[t]{0.4\columnwidth}Apple $\rightarrow$ Orange, Guava, Pillow\end{minipage} \tabularnewline
    \hline 
    \end{tabular}
    }
    \caption{\footnotesize{\textbf{Generalization test set} The different types of test-scenarios that constitute to the Generalization data set (GenTest) consisting of a total of $1,406$ cases. The test scenarios sample novel spatial contexts and goals, perturb 
    human demonstrated tools to alternative relevant and non-relevant tools and change the goal 
    characteristics. }}
    \label{tab:generalization_testcases}
\end{table}


\begin{table}
    \centering
    \resizebox{\columnwidth}{!}{
    \begin{tabular}{|c|c|c|c|c|c|c|c|c|c|}
    \hline 
    \multirow{2}{*}{Model} & \multicolumn{2}{c|}{Test} & \multicolumn{5}{c|}{GenTest-Type} & \multicolumn{2}{c|}{GenTest}\tabularnewline
    \cline{2-10} 
     & Home & Factory & I & II & III & IV & V & Home & Factory\tabularnewline
    \hline 
    \hline 
    GGCN Baseline &  61.53 &  86.36 & 30.15 & 20.07 & 23.51 & 35.79 & 56.93 &  55.93 &  23.08\tabularnewline
    \hline 
    \hline 
    +Metric & 72.30 & 89.77 & 66.67 & 43.31 & 54.34 & 35.79 & 70.53 & 60.08 & 46.04\tabularnewline
    \hline 
    +Attn & 83.07 & 91.06 & 66.67 & 73.67 & 61.28 & 40.34 & 93.56 & 63.05 & 67.75\tabularnewline
    \hline 
    +L & 70.76 & 89.77 & 66.67 & 80.97 & 75 & 50 & 97.34 & 67.21 & 72.54\tabularnewline
    \hline 
    +NT & 83.07 & 95.12 & 66.67 & \textbf{100} & 75 & 50 & 97.34 & 82.19 & 72.54\tabularnewline
    \hline 
    +C & \textbf{88.88} & \textbf{100} & 71.43 & \textbf{100} & \textbf{100} & 59.21 & \textbf{100} & 91.09 & 88.66\tabularnewline
    \hline 
    +W (\modelname) & \textbf{88.88} & \textbf{100} & \textbf{100} & \textbf{100} & \textbf{100} & \textbf{64.56} & \textbf{100} & \textbf{100} & \textbf{90.02}\tabularnewline
    \hline 
    \end{tabular} }
    \caption{\footnotesize{\textbf{Evaluation:} Model prediction accuracies on the Test and the GenTest for 
    the Home and Factory Domains. The accuracies are reported as per the incremental addition 
    of model components. The model shows improvements in test and generalization accuracies over the 
    baseline model from \citet{liao2019synthesizing}. }}
    \label{tab:comparison}
\end{table}

\begin{table*}
    \centering
    \resizebox{\textwidth}{!}{
    \begin{tabular}{|c|c|c|c|c|}
    \hline 
    \multicolumn{5}{|c|}{Home: Place fruits in cupboard generalization test case (\emph{tray} unavailable)}\tabularnewline
    \hline 
    \includegraphics[width=0.195\textwidth]{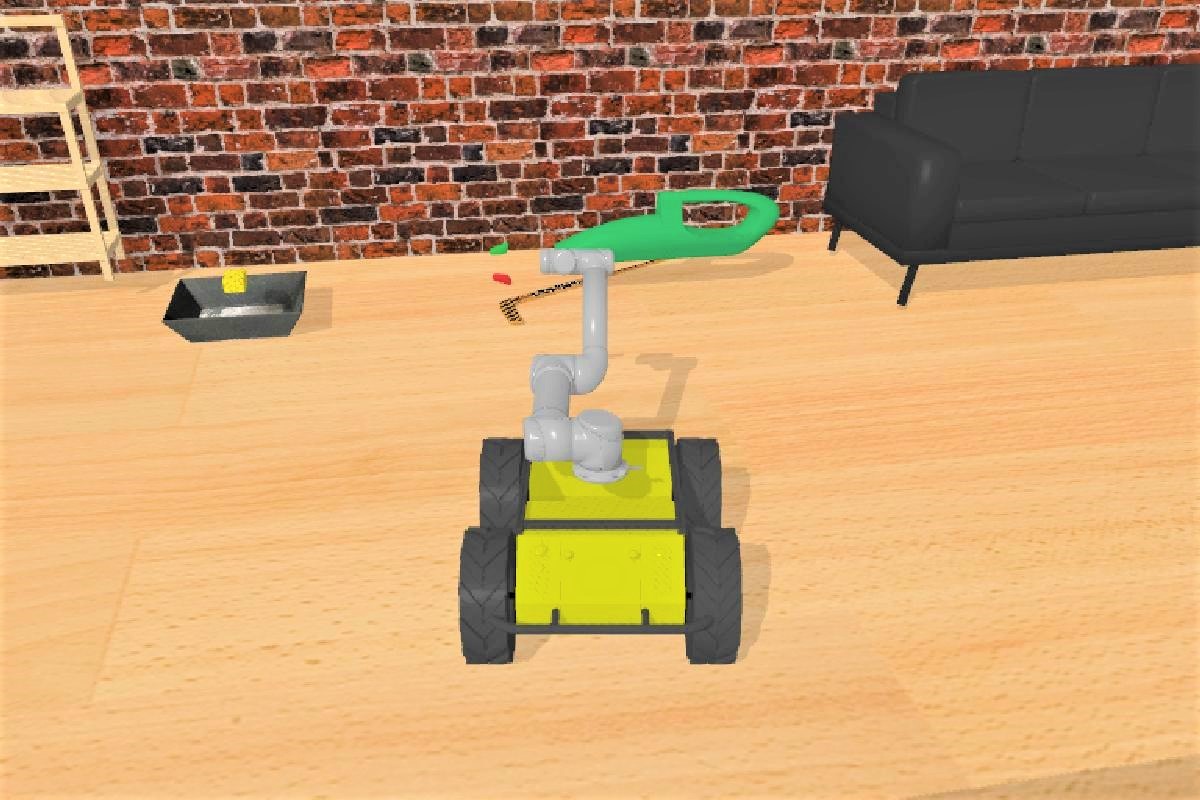} & \includegraphics[width=0.195\textwidth]{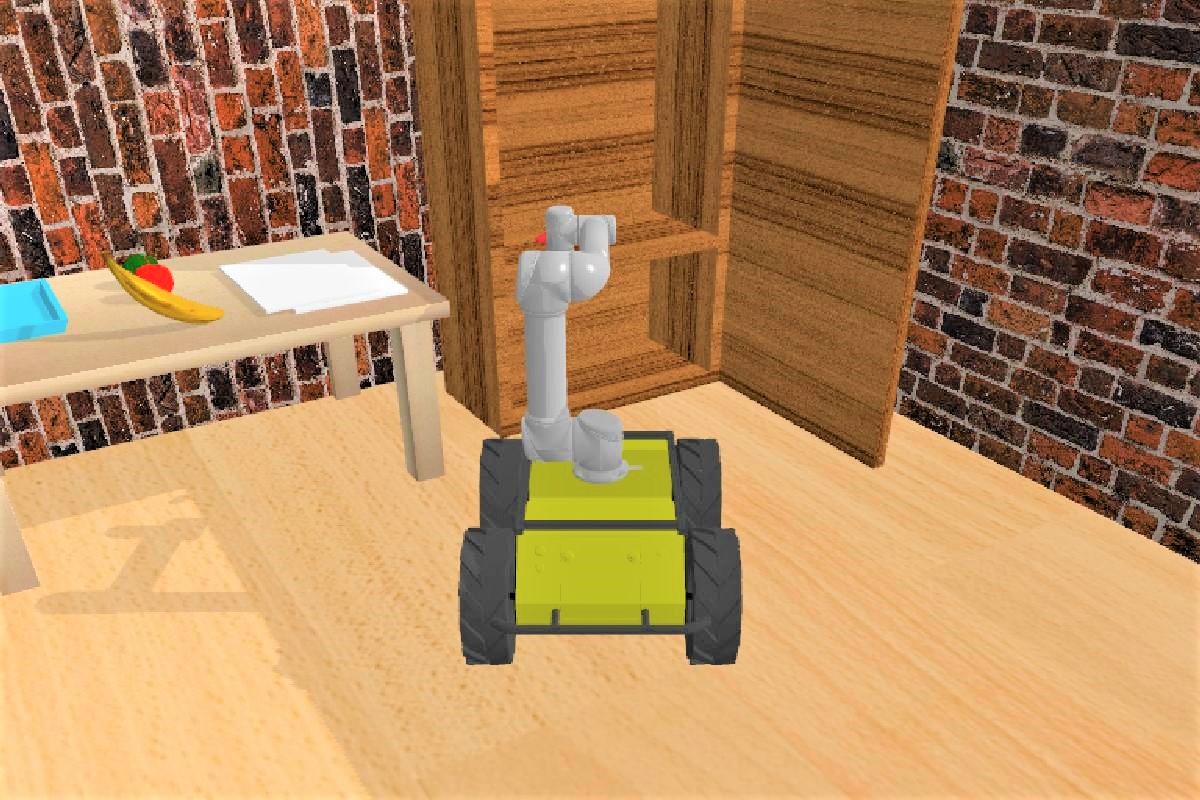} & \includegraphics[width=0.195\textwidth]{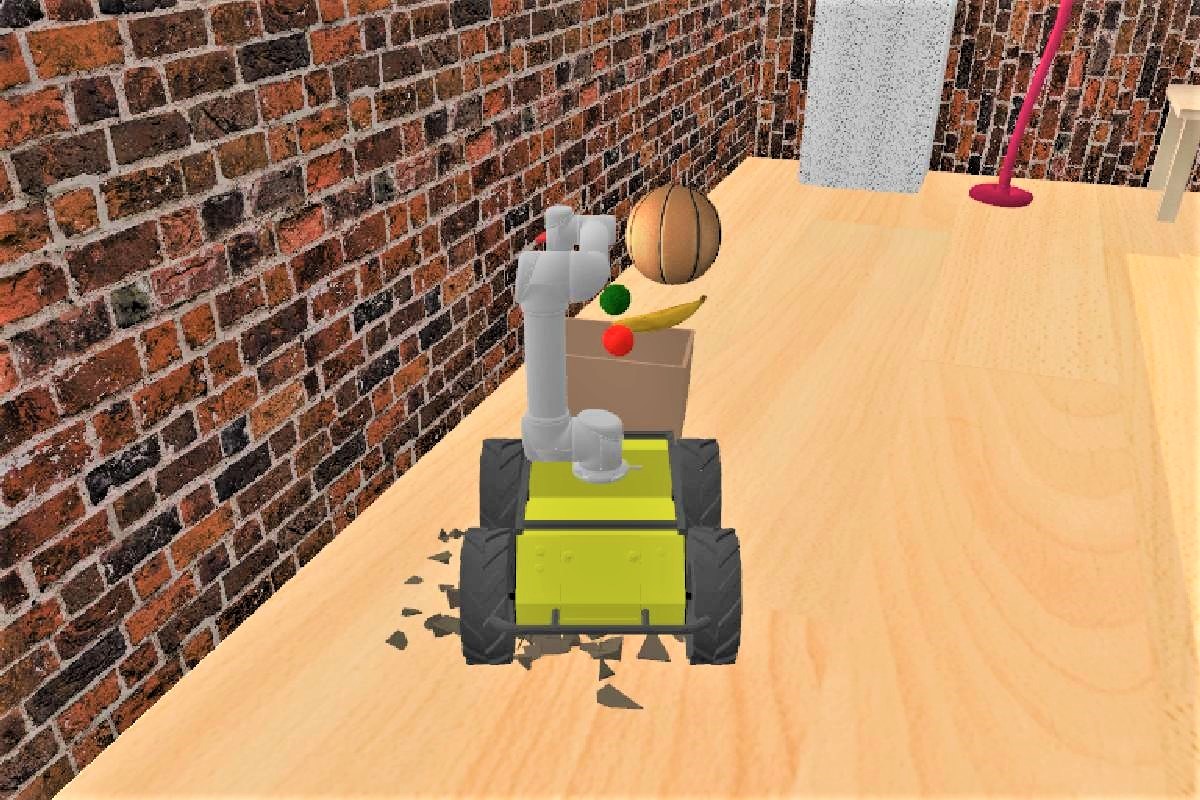} & \includegraphics[width=0.195\textwidth]{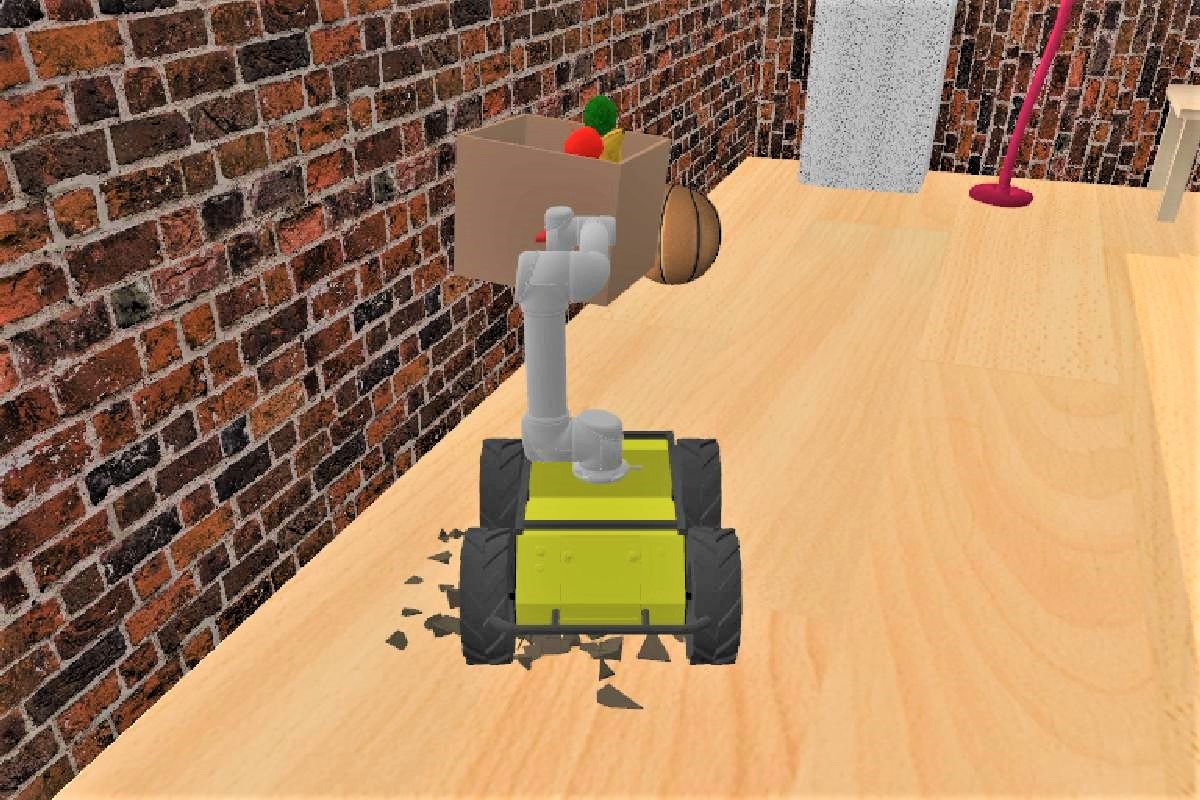} & \includegraphics[width=0.195\textwidth]{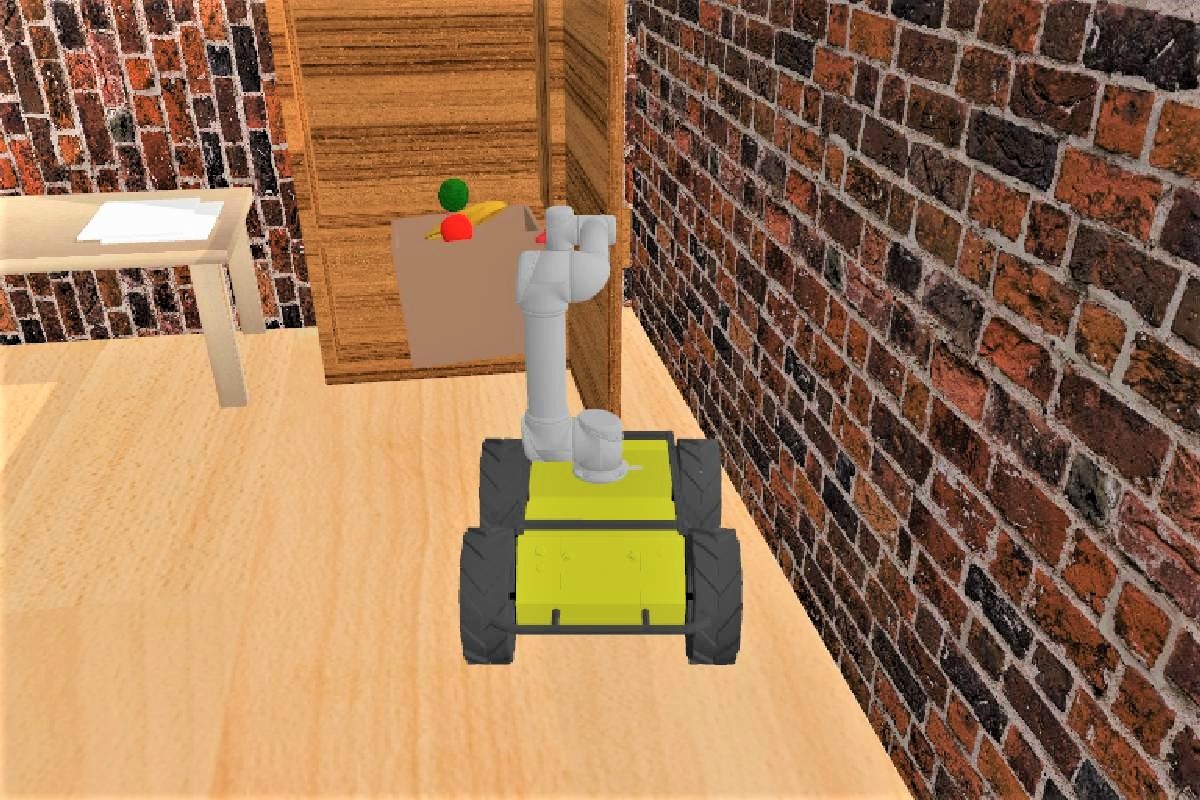}\tabularnewline
    \hline 
    0: Predict Box & 29: Open cupboard & 204: Place fruits in box & 216: Pick box & 243: Place box in cupboard\tabularnewline
    \hline 
    \multicolumn{5}{|c|}{Factory: Fix board on wall generalization test case (\emph{screws} unavailable)}\tabularnewline
    \hline 
    \includegraphics[width=0.195\textwidth]{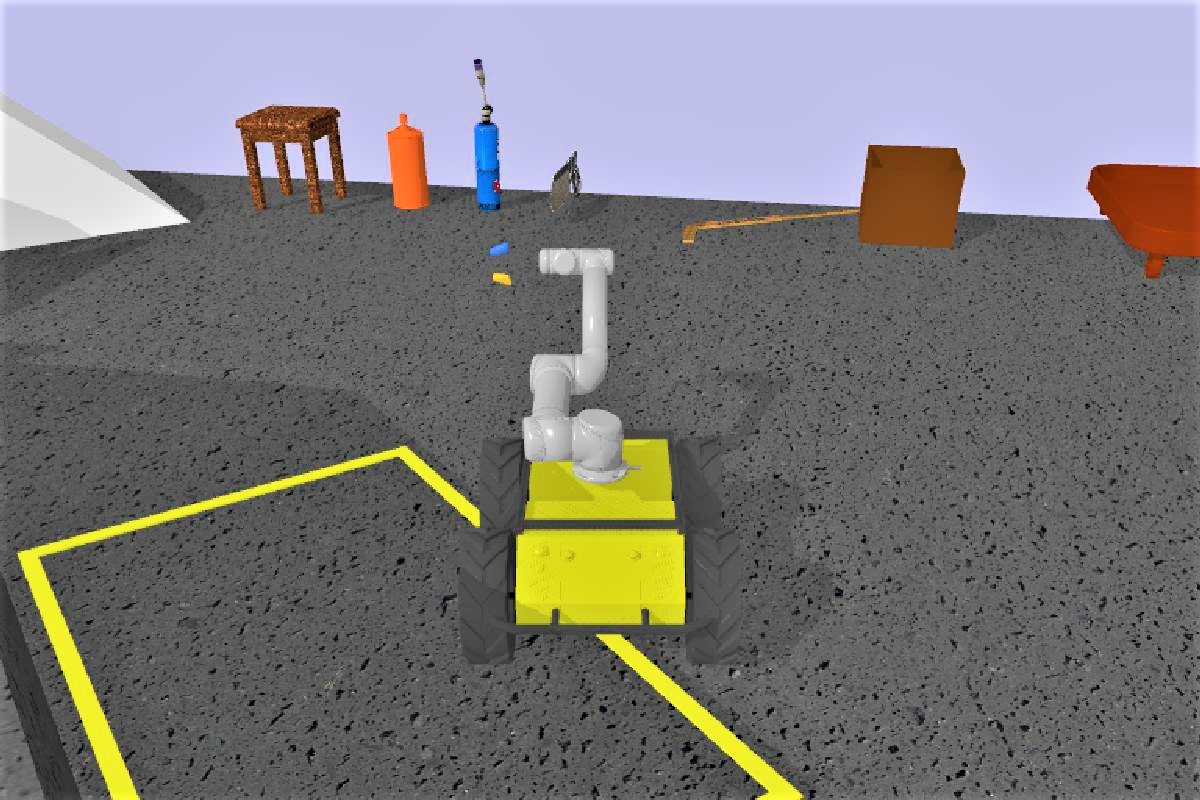} & \includegraphics[width=0.195\textwidth]{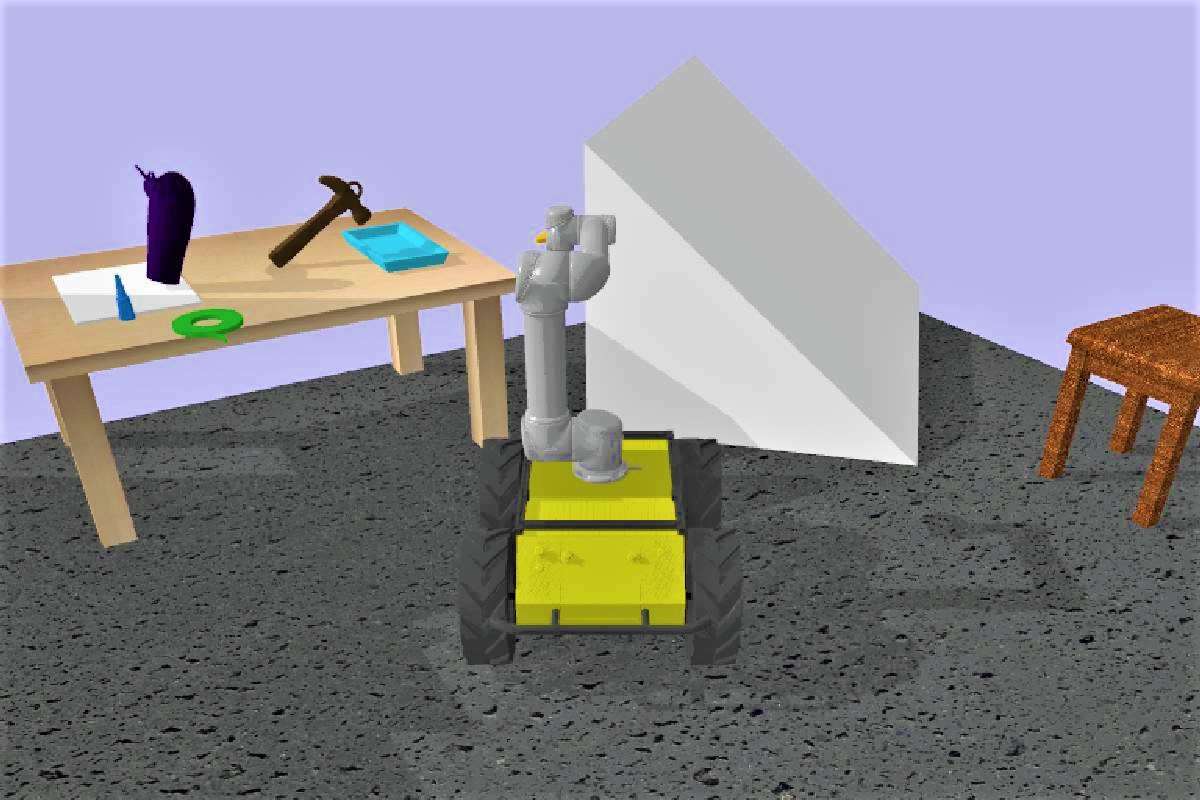} & \includegraphics[width=0.195\textwidth]{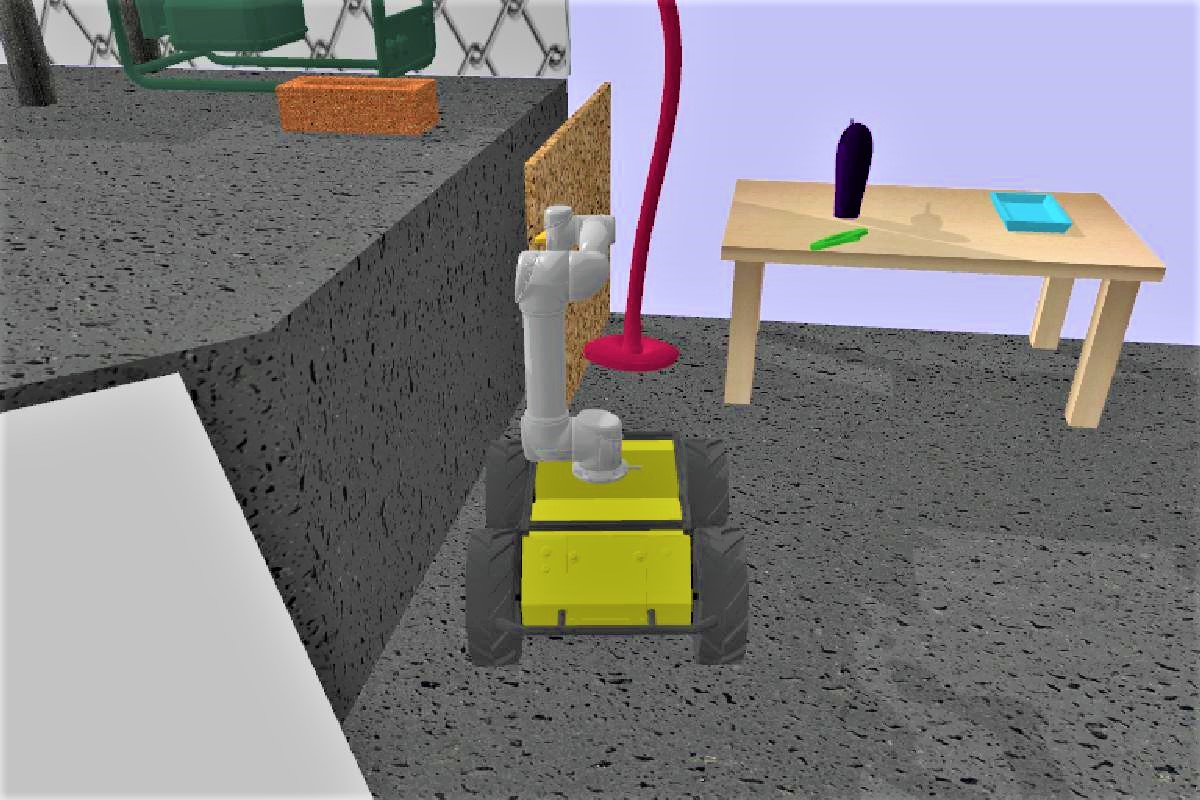} & \includegraphics[width=0.195\textwidth]{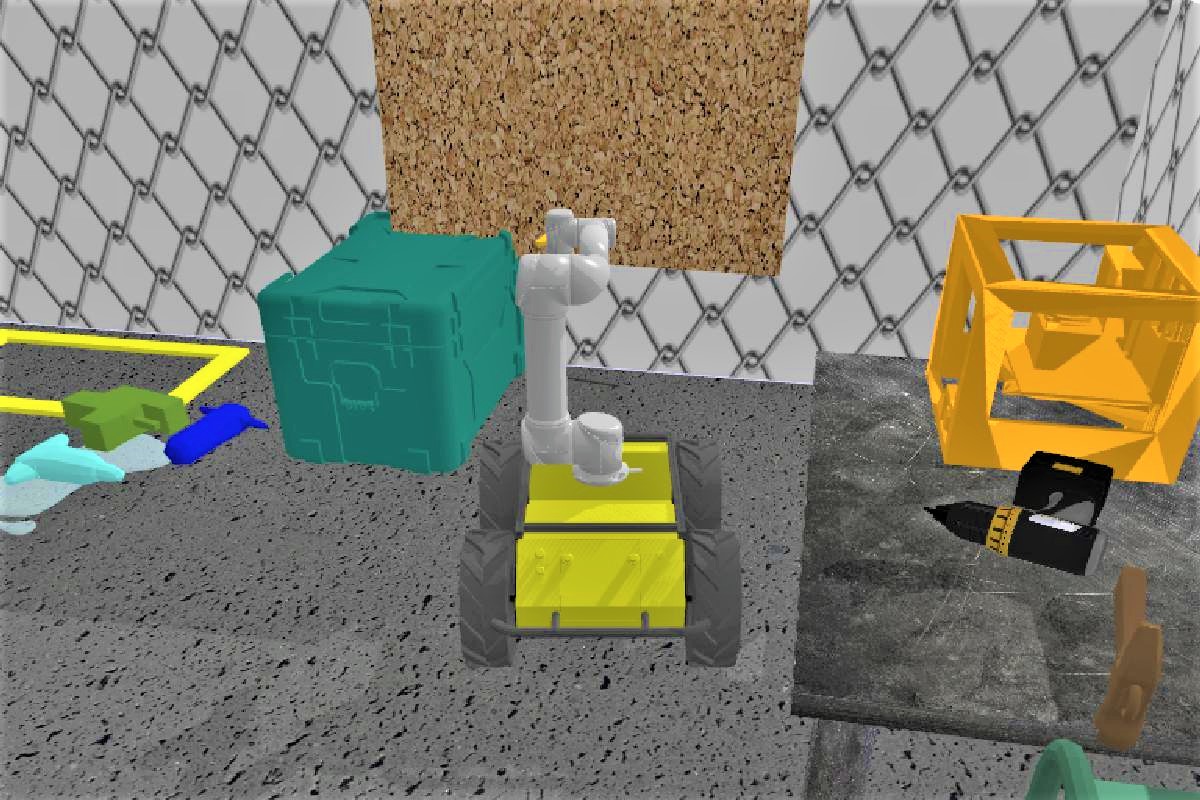} & \includegraphics[width=0.195\textwidth]{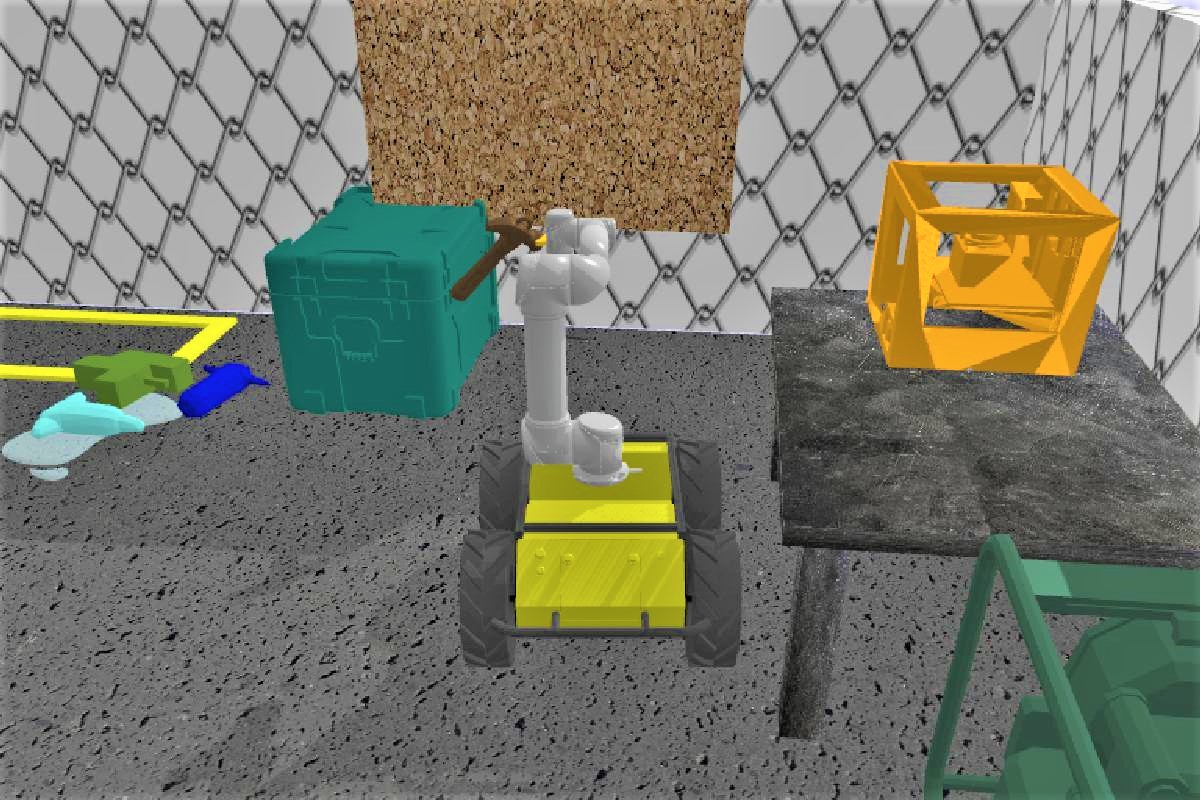}\tabularnewline
    \hline 
    0: Predict hammer & 21: Place ramp & 80: Pick Board & 177: Place board on wall & 214: Hammer nail on board\tabularnewline
    \hline 
    \end{tabular}
    }
    \caption{
    \footnotesize{\textbf{Utility of model predictions. } Preliminary experiments indicate 
    the utility of model predictions in synthesizing a feasible symbolic plans 
    for a simulated mobile manipulator in home and factory domains.   
    The learned model predicts the utility of tool objects a-priori unseen in training. 
    The model prediction likelihoods are used to inform exploration of 
    robot interactions (including tool use) during plan search. 
    Figure illustrates the robot using novel tools to complete the task (images in 
    temporal order with execution time indicated above). }
     %
     %
    }
    \label{tab:plans}
\end{table*}

Table~\ref{tab:comparison} shows the final accuracies on the test-set (Test) and generalization test-set (GenTest) 
with individual accuracies for each test-type. The results of the complete  \modelname architecture are in the last row of the table. On the regular test set,  \modelname outperforms the GGCN baseline by 14 and 27 accuracy points on Home and Factory domains, respectively. Each model component improves the accuracy numbers, with the exception of factored likelihood -- it makes the model more complex to aid prediction of an unseen tool as the output. In all earlier models, each tool is an independent class; that restriction solves an easier problem obtaining better performance on the regular test set.

A similar pattern is found in the generalization test set, where each model component brings tremendous value. The improvement in accuracy is dramatic in Factory domain: a 67 point accuracy improvement is seen on top of the GGCN baseline. 
Analyzing a component's effectiveness across different generalization types, we obtain further insights on models' workings. The late fusion of metric properties of each object (+Metric) allows such information to be emphasized in model. This gets significant improvements in almost all test cases. For instance, it allows the model to predict \emph{box} (instead of \emph{tray}) when transporting a pillow by using pillow's size information (a metric property). Major improvements are observed in Type I as those scenarios require reasoning about object nearness based on their locations (another metric property).

%

%

The \emph{Attn} component also makes across the board improvements with maximum impact to types II and V. The impact to Type V is natural, since goal objects get replaced in those examples. Explicitly biasing the model to use the features of those objects (through conditioned attention) increases their importance, and likely reduces overfitting. An example for Type II is when \emph{generator} is specified in the goal, and wood (the fuel for generator, and the most likely tool) is made absent from the scene. The model could err in  giving attention to the \emph{wood-cutter} tool, which is often correlated with wood. However, conditioned attention gives low attention to \emph{wood-cutter} and predicts \emph{gasoline}, instead.

%
%
The factored likelihood (+\emph{L}) predictably helps the most in Type III scenarios, since without this component, the model cannot predict any unseen tool. A decent performance of earlier models on Type III is attributed to alternative possible correct answers (any alternative seen tool or no-tool) due to multiple annotations per scenario.
%
The NT component, which splits the problem into two predictions (whether to use the tool and which one), helps in Type II cases, where the most likely tool is removed. In such cases a \emph{no-tool} prediction is often correct, which is correctly predicted by the NT predictor focusing on whether to use the tool.
%
The $\mathrm{ConceptNet}$ embeddings (+C) likely contain commonsense knowledge about unseen tools and objects, for example, whether a new tool is flat or not (which should help in ascertaining whether it can be used for transport or not). Using these embeddings makes huge improvement in Type III cases where entirely new objects are to be predicted as tools. 
%
%
Finally, giving higher weight to optimal plans (+\emph{W}) allows the model to differentiate tools by plan execution time and not human usage frequency.
This helps in improved metric generalization, predicting nearby tools in test-type I. Overall, the complete architecture provides the maximum generalization accuracy among all models.

Additionally, we performed preliminary experiments to assess the utility of the learned 
model in aiding plan synthesis. 
We used an symbolic planner that encoded the world state and robot action representation 
described previously. The planner used uninformed search to explore the space of 
robot interactions in the environment during plan search.   
The estimated tool likelihoods 
prioritized the exploration of candidate object interactions while 
expanding the search tree. 
Table~\ref{tab:plans} illustrates cases where the model predicts the use of 
novel tools: a ``box" object for transport task and a ``nails" to attach a board to the wall. An underlying symbolic planner prioritizes model predictions over other possible 
tool object interactions during search for a feasible plan. 
Preliminary evaluation revealed a reduction in the effective branching factor 
from $37.29\pm1.99$ to $4.24\pm0.13$ in the Home domain and $56.89\pm4.01$ to $9.10\pm3.21$ in the Factory domain using the 
predictions of the learned model compared to an uninformed plan search. 
However, extensive evaluation remains part of future work.

\section{Conclusions} \label{sec:conclusions}

We addressed the problem of learning common sense knowledge 
of contextual tool use for a robot operating in environments 
with potentially new objects not encountered before. 
We crowd source a data set of robot plans where a humans instructs
 a simulated robot to perform tasks involving interaction 
 with objects as tools in home and factory-like environments. 
The demonstrated plans are used to train \modelname, a neural learner that predicts the contextual use of tools enabling the robot to synthesize a plan for the intended goal. 
The model builds on gated graph convolution networks and 
incorporates goal-conditioned attention, fusion of semantic and metric 
representations and use of existing knowledge sources such as ConceptNet. 
The imitation learner demonstrates accurate generalization to environments with novel object instances using the learned knowledge of shared spatial and 
semantic characteristics. 

Future work will investigate use of the learned model with a symbolic planner, 
handling partially observable environments 
and extensions for imitating multi-step tool interactions. 
\section*{Acknowledgments}
Rohan Paul is supported by the Pankaj Gupta Faculty Fellowship and DST’s Technology Innovation Hub (TIH) for Cobotics. Mausam is supported by an IBM SUR award, grants by Google, Bloomberg and 1MG, Jai Gupta chair fellowship, and a Visvesvaraya faculty award by Govt. of India. We thank \citet{puig2018virtualhome} for sharing the implementation for baseline comparison.  We thank Keshav and Anil Sharma for their assistance in setting up the data collection  infrastructure.  We are grateful to anonymous turkers and student volunteers for assisting in the data collection.  We thank Pulkit Sapra and Prof. P. V. M. Rao for assistance with CAD model creation. We thank Jigyasa Gupta for comments on improving the paper presentation. We thank IIT Delhi HPC facility for compute resources.

\bibliographystyle{unsrtnat}
\bibliography{references}

\end{document}